# Explanation of Reinforcement Learning Model in Dynamic Multi-Agent System

Xinzhi Wang*
School of Computer Engineering and Science
Shanghai University
Shanghai, 200444, China

Huao Li†
School of Computing and Information
University of Pittsburgh
Pittsburgh, PA 15260, USA

Hui Zhang
Institute of Public Safety Research Tsinghua University
Beijing, 100084, China

Michael Lewis
School of Computing and Information
University of Pittsburgh
Pittsburgh, PA 15260, USA

Katia Sycara
Robotics Institute Carnegie Mellon University
Pittsburgh, PA 15213, USA

## ABSTRACT

Recently, there has been increasing interest in transparency and interpretability in Deep Reinforcement Learning (DRL) systems. Verbal explanations, as the most natural way of communication in our daily life, deserve more attention, since they allow users to gain a better understanding of the system which ultimately could lead to a high level of trust and smooth collaboration. This paper reports a novel work in generating verbal explanations for DRL agent's behaviors. A rule-based model is designed to construct explanations using a series of rules which are predefined with prior knowledge. A learning model is then proposed to expand the implicit logic of generating verbal explanation to general situations by employing rule-based explanations as training data. The learning model is shown to have better flexibility and generalizability than the static rule-based model. The performance of both models is evaluated quantitatively through objective metrics. The results show that verbal explanation generated by both models improve users' subjective satisfaction towards the interpretability of DRL systems. Additionally, seven variants of the learning model are designed to illustrate the contribution of input channels, attention mechanism, and proposed encoder in improving the quality of verbal explanation.

## 1 INTRODUCTION

Deep Reinforcement learning (DRL) has been applied in multiple domains including object detection [21], robotics control [14, 20] and natural language processing [17]. The DRL networks are also successful in playing games such as Atari games [22] and Go [26]. However, the inner logic and reasoning of DRL systems are opaque and difficult to be understood even by their own designers. The human factors and computational literature [12, 19, 23] has pointed out the need for system transparency as a way to increase trust in the system. Additionally, transparency would be useful for human collaboration with autonomous agents that use DRL. When interacting with autonomous intelligent agents, people tend to regard them as intentional individuals and explain their behaviors as in interpersonal relationships [2]. That requires an explainable agent to clarify its actions by offering reasons of beliefs, desires, and intentions [15]. Additionally, system transparency can provide clues for designers to debug system.

While some amount of work has been done in computer vision to make Deep Neural Networks more transparent [3, 13, 16], there are only a couple of works [17], [11], [7] that address transparency in DRL Networks. This is due to the additional complexity of learning for sequential decision making. Researchers [11] found that visualizations of DRL reasoning via a new technique, object-saliencey maps, were as effective in enabling subjects to make predictions about a DRL agent's (Ms. Pacman) next actions as giving the subjects access to sequences of game screenshots. Although these results were promising, the human's predictions had accuracy of 60%. One of the possible reasons was that in the object-saliencey maps there were multiple objects that Ms. Pacman could attend to and that could influence its subsequent decisions. This could induce ambiguity in the subjects' mind as to which action Pacman would take next. We, therefore, decided to develop focused verbal explanation models of the DRL system since (a) the verbal explanation could refer to objects that would be the most important in influencing the agent's selection of next action and (b) language may be more satisfying to people as a means of communication.

However, no public datasets on verbal explanation of DRL systems are available currently. Therefore we chose to construct explanation datasets and verbal explanation models by ourselves. This paper reports on the methods of generating the datasets and models.

Since the verbal explanation of DRL system pertains to explaining the internal logic of the agent, rahter than a human subjective

*Xinzhi Wang is the corresponding author.
†Xinzhi Wang and Huao Li contribute equally to this work.

interpetation on the agent's behavior, we generated an initial rule-based model based on prior knowledge of the Ms. Pacman game and its rules. Although the rule-based model is capable of generating reasonable explanations, it lacks of generalizability and flexibility for unexpected situations. In these cases, neural network based learning models are entrusted with the responsibility. They can learn the implicit logic of generating verbal explanation for DRL system through neural connections, once given sufficient training data. The challenge for the learning model lies in extracting distinguishable features from DRL systems, especially from high structural similar images, such as game image with fixed board or fixed map, which can not be fully solved by existing networks currently.

In this paper, both rule-based model and learning model are presented to generate verbal explanation for DRL systems. Game image, agent position map, and object saliency map act as the input for both models. Data generated by the rule-based model is employed to train the learning model, which consists of two parts: an encoder on feature extraction, and a decoder on generating the explanation in natural language with attention mechanism. We make the following contributions:

- To reduce the ambiguity of saliency maps and to improve the rationalization of explanation, we put forward conceptual level verbal explanation employing natural language and test human subject satisfaction.
- To generate objective verbal explanation for DRL systems, both rule-based model and learning model are introduced. In the rule-based model, a group of rules aiming at the position, relation, and property of agents are designed and evaluated. In the learning model, a new convolutional neural network and attention mechanism is put forward to extract distinguishable features from structural images.
- The performance of the model on Atari game Ms. Pacman is quantitatively validated. The results show the learning model can generate verbal explanations with high quality, which would contribute to improved human understanding of DRL systems.
- Finally, this paper aims to set clues for explaining DRL systems objectively, combining both the rule-based model and the learning model. We employ the Atari game Ms. Pacman as example of the application scenario to build rules through prior knowledge. The rules need to be rebuilt and the learning model need to be retrained for new applications.

Figure 1 provides a conceptual overview of the roles of the rule based-model and learning model using a screenshot and the corresponding o-saliencey map from Ms. Pacman.

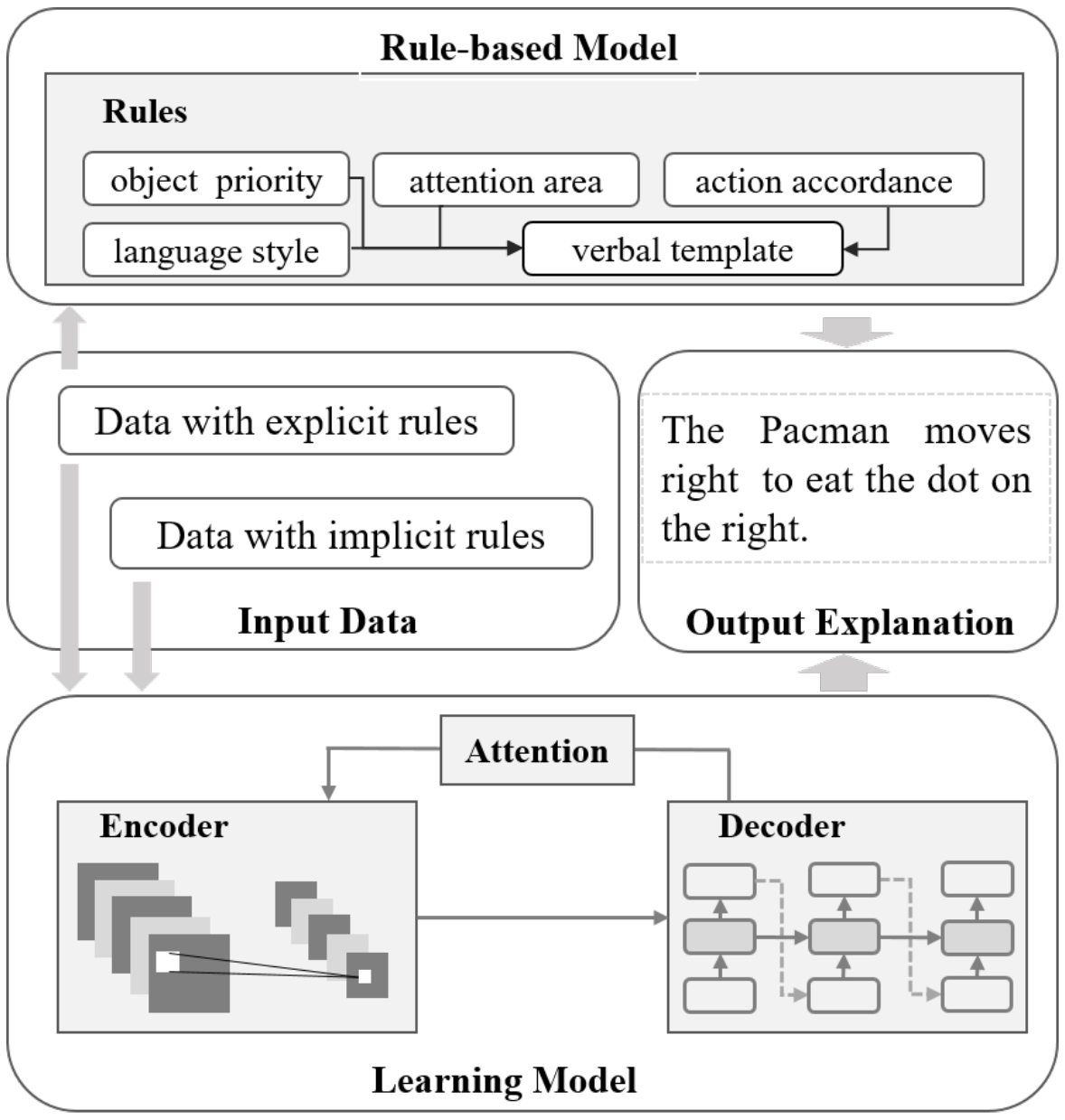

**Figure 1: The rule-based model supports learning model by providing training data. The learning model generalizes the rule-based model to all the situations.**

## 2 RELATED WORK

The works of [11, 17, 31] are the most relevant to this paper since they are the ones that address transparency in DRL agents. They generate object saliency maps for visual interpretation and explanation of DRL agent's decisions. They also evaluate the usefulness of the visual interpretation via human experimenation. [7] uses pixel saliency to focus mainly on situations of agent overfitting to provide insight for designers and they do not provide any human evaluation.

Less relevant work is image classification and object recognition in the literature [16], and extracting the most discriminating properties of visible object [10]. Some works explain models through images. Koh et al. [13] designed influence functions to trace the most responsible training points for a given prediction. Du et al. [3] proposed a guided feature inversion framework. Erhan et al. [5] visualized deep models by finding an input image which maximizes the neuron activity of interest by carrying out an optimization using gradient ascent in the image space. The same method was later employed by [16] to visualize the class models, which captured by a deep unsupervised auto-encoder. Simonyan et al. [27] proposed pixel saliency maps to deduce the spatial support of a particular class in a given image based on the derivative of class score with respect to the input image. Ribeiro et al. [23] proposed a method to explain the prediction of any classifier by local exploration, and apply it on image and text classification. Such works provide important clues to the transparency of DRL system. Lots of these works, though initially proposed for interpreting image classifiers, can be transferred to explain DRL system, such as vanilla pixel saliency [27], guided backpropogation [29], GradCAM [24], and mask based optimization approach [6].

There are also natural language works such as interactive robot dialogue system [25], image caption generation [30], image caption with attention [32]. The models, which transfer image into language, are valuable for designing our model. The following works specialized in generating verbal explanations. De Graaf and Malle [2] studied the explanation in autonomous intelligent systems through the way of people explaining human behavior. Hayes et al. [8] presented algorithms which enable robots to synthesize policy descriptions. Das et al.[1] posed a cooperative game between two agents who communicate in natural language based on an unseen image from a lineup of images by employing deep reinforcement learning (RL) to learn the dialog policies. Shridhar et al.[25] presented a robot system that follows human natural language instructions to pick and place everyday objects through interactively asking questions to disambiguate referring expressions. Ehsan et al.[4] described a technique which translates internal state-action representations of autonomous agent into natural language. However, they either lack of generalization, or involve human subjectivity. We proposed a generalized verbal generation model with little human subjective interference in this paper with the support of all the previous works.

# 3 RULE-BASED VERBAL EXPLANATION MODEL

In this section, the rule-based model is introduced. We take the Atari game Ms. Pacman as instance to define the rules in this section. The input of rule-based model is game images and corresponding object saliency maps. The output is verbal explanation results of the given game state.

## 3.1 Object Saliency Map

Object saliency map provides potential transparency by revealing the importance of each object [11, 17]. They rank the objects in a state $s$ based on their effect on Q-value $Q(s, \alpha)$ by masking the object with background color to form a new state $s_o$ as if the object does not appear in this new state. Then calculate the Q-values for the states with and without the object, making the difference of the Q-values $Q(s_o, \alpha) - Q(s, \alpha)$ to represent the object influence on $Q(s, \alpha)$.

## 3.2 Criterion of objects

In the game Ms. Pacman, objects except the Pacman herself can be divided into two categories. One is objects that the Pacman can eat to accumulate points, including edible ghosts, cherries, pellets, and dots. The other is objects that the Pacman needs to avoid to stay alive which only refer to ghosts. According to the function of objects, the expected $Q(s_o, \alpha) - Q(s, \alpha)$ of different objects varies as following:

- For cherries, pellets, dots and edible ghosts, the $Q(s_o, \alpha) - Q(s, \alpha)$ should be negative.
- For ghosts, the $Q(s_o, \alpha) - Q(s, \alpha)$ should be positive.

## 3.3 Rules definition

Based on the understanding of game rules and human-generated explanations from previous study [11], rules are designed for objects and actions respectively. The rules collaborate with each other to generate the verbal explanations.

### 3.3.1 *Object priority.*

Considering the different weights of objects in the rewarding point, we represent different objects in order and selectively when generating the explanation. The priority of objects is ghost, edible ghost, cherry, pellet, and dot.

### 3.3.2 *Attention area.*

It's beneficial for a explanation to contain minimal valuable information from the psychological perspective [18]. Taking that into account, an attention area is designed to only consider objects within a certain distance from the Pacman. Also, convincing explanations should contain concise information [2]. As a result, we limit the maximum number of objects appear in the verbal explanation, to four.

### 3.3.3 *Action accordance.*

The expected action of Pacman should be approaching beneficial objects and avoiding ghost. However, there are situations where Pacman has to leave beneficial object in order to avoid ghost or Pacman has to approach ghost in order to chase beneficial objects. The action of Pacman is divided into two classes, considering consistent or inconsistent with expectations. Class #1 means the action of Pacman is in accordance with the expectation. Class #2 means the action is in contrast with the expectation.

### 3.3.4 *Language style.*

We choose the language style by following human's habit, which makes the explanation process naturally and fluently. Moving direction is first introduced. Relative coordination is employ to indicate the position of objects to Pacman.

All the former rules collaborate together to generate verbal explanation for DRL systems.

## 3.4 Workflow

To generate verbal explanation employing the former designed rules, the workflow is proposed.

I. Unified the Q-value of objects. Q-value of edible ghost, cherry, pellet, dot are multiplied by -1. In this way, the higher the better for all objects.
II. Set the attention area.
III. Filter the special objects (Ghost, Edible ghost, Cherry, Pellet) by the Q-value. Four objects with highest positive Q-value are kept. At most one dot is kept who carry the highest Q-value of all dots. and the Q-value is higher than the threshold (lowest) of the four (at most) special objects.
IV. Classify objects into two classes based on the action accordance.
   a. If Class #1 and Class #2 appears simultaneously. Explain Class #1 (as the reason) and then explain Class #2 (as the result)
   b. If only Class #1 appears, then explain Class #1
   c. If only Class #2 appears or neither Class #1 and Class #2 appears, then, unexplainable. Skip.
V. Generate verbal explanation employing pre-defined template.

# 4 LEARNING-BASED VERBAL EXPLANATION MODEL

The rule-based model is able to generate verbal explanations, but lacks of generalizability and flexibility for unexpected situations as mentioned in the introduction. The performance of the rule-based model becomes unreliable if the rules don't apply for the particular episode or the saliency map fails due to recognizer problems etc.. To overcome the drawbacks, the learning model is proposed which advances in the following parts: 1) It is more generalizable in terms of game episodes 2) It is more tolerant to input noises 3) It is able to provide explanation for the future timestamps, which help user predict and plan future states ahead.

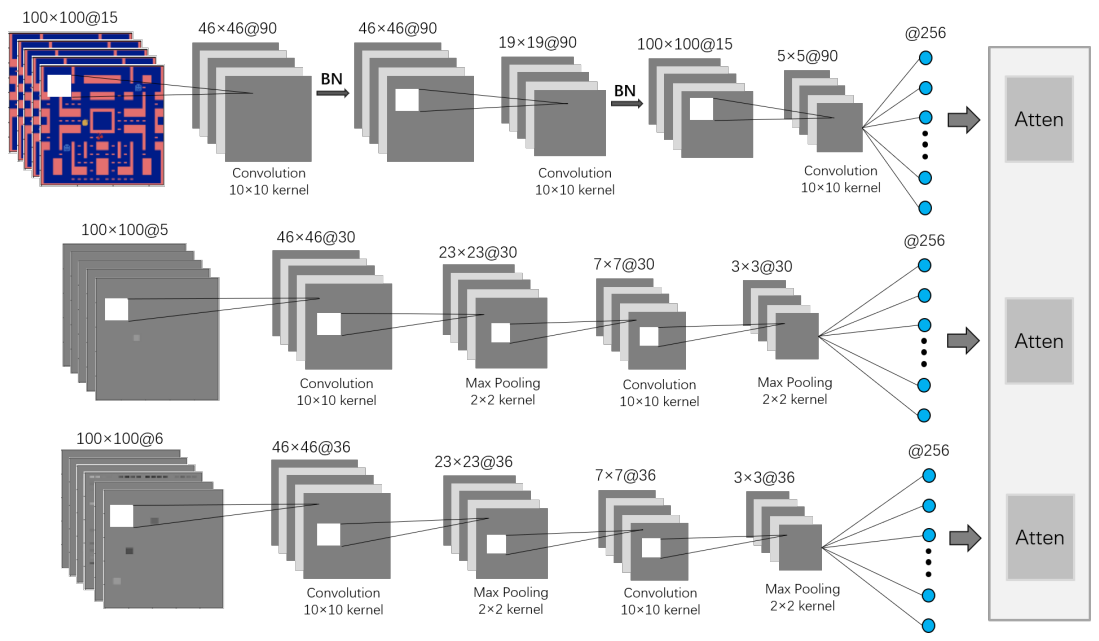

**Figure 2: The image encoder processing network**

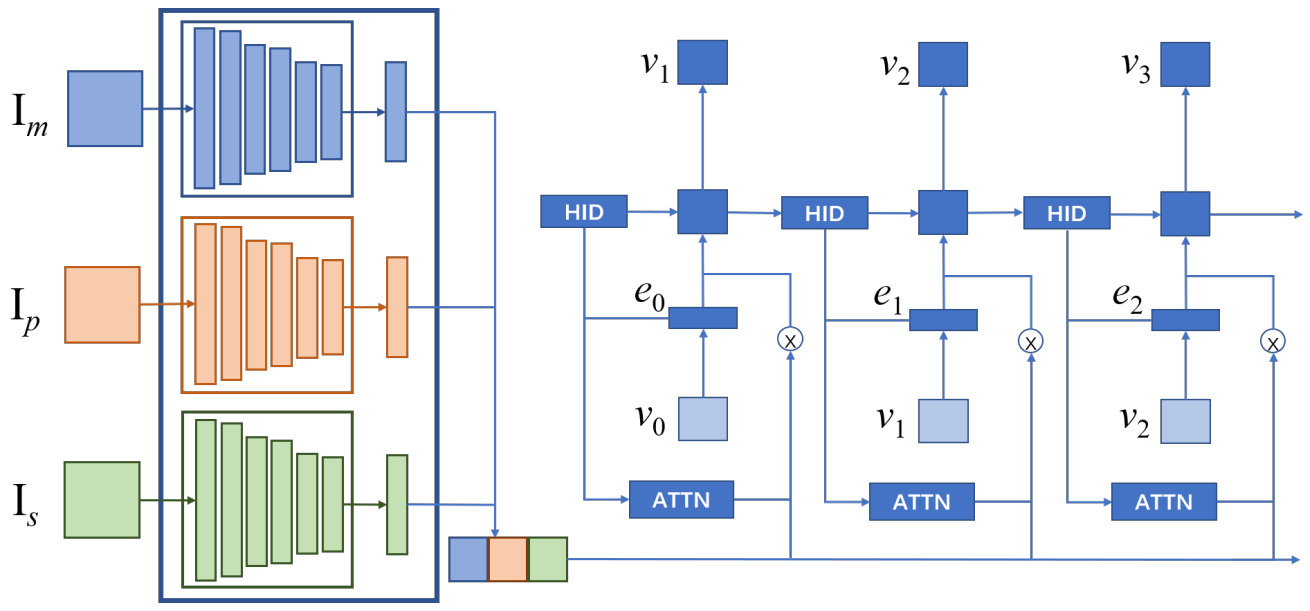

**Figure 3: Learning model for verbal generation with self-designed encoder and attention on three kinds of input.**

The learning model consists of two stages. The first stage is image processing to get game image feature. The second stage is to generate verbal content based on the image process result. To capture dynamic game environment information, and to get the impact of object to Pacman, five previous frames of game image, five previous frames of Pacman position map, and a frame of object saliency map are selected as input. The verbal description of game image is generated as output. The overall working process in shown in figure 3.

If we use the image input of time step $(t-1)$ to generate verbal explanation of time step $t$ in game episode, then the model becomes a prediction model. The saliency map in this paper is object saliency map, which is proved to be potential in previous work [11].

## 4.1 Image Encoder

Image encoder aims to encode game image, Pacman position map, and object saliency map for the verbal decoder part. Game image provides environment information. Pacman position map carry Pacman location information and object saliency map provide weight information of each object. The encoder extracts potential features, reveals the inherit causal relations among objects, and explores the visual relations. The details are described as follows.

Let $\mathrm{I}_m \in \mathbb{R}^{100\times100\times15}$ being the matrix of game image, which contains fifteen channels, including five consequent of game images and each frame with three channels (RGB). $\mathrm{I}_p \in \mathbb{R}^{100\times100\times5}$ is the pacman position map with five consequent frames and each frame with one channel. $\mathrm{I}_s \in \mathbb{R}^{100\times100\times6}$ indicates matrix of the object saliency maps extracted from the last frame of five game image frames. The six channels are object saliency maps for ghost, edible ghost, cherry, pellet, dot, and Pacman.The structure of encoder for game image, Pacman position map, and object saliency map, is shown as follows:

$$\mathrm{I}_m' = f_m(\mathrm{I}_m|\mathrm{W}_m, \mathrm{B}_m) \tag{1}$$

where $\mathrm{I}_m'$ is the state processing network result. $\mathrm{W}_m$ and $\mathrm{B}_m$ are parameters for the calculation. Function $f_m$ includes two convolution calculation with two $10\times10$ kernels whose stride is set to be 1, two Relu activation operation following the convolution calculation and two batch normalization.

$$\mathrm{I}_p' = f_p(\mathrm{I}_p|\mathrm{W}_p, \mathrm{B}_p) \tag{2}$$

where $\mathrm{I}_p'$ is the pacman processing network result. $\mathrm{W}_p$ and $\mathrm{B}_p$ are parameters for function $f_p$. The difference between $f_m$ and $f_p$ is that $f_p$ substitute the two batch normalization operation with two max pooling operation whose kernels are set to be $2\times2$.

$$\mathrm{I}_s' = f_s(\mathrm{I}_s|\mathrm{W}_s, \mathrm{B}_s) \tag{3}$$

where $\mathrm{I}_s'$ is the saliency map processing network result. Function $f_s$ has the same structure with $f_p$ but different parameters.

## 4.2 Verbal Decoder

Verbal decoder aims to generate verbal description about the image encoder output. Sequence generation model is employed to generate verbal explanation verbatim. Attention mechanism on three outputs from encoder is designed to select the most salient output. Before we go deep in the language generation part, we define three operations, where $\eta(\cdot)$ is softmax operation, $\sigma(\cdot)$ means sigmoid activation, and $\rho(\cdot)$ indicates ReLU activation.

Let $\mathrm{V} = \{v_1, ..., v_N\}$ be the verbal explanation of game image, where $N$ is the number of words in the verbal explanation, $v_i \in \{0,1\}^N$ is the one-hot representation of corresponding word. $e_i \in \mathbb{R}^{D_e}$ is the embedding vector for the $v_i$, $D_e$ is the dimension of the word embedding. Also, let $a_t \in \mathbb{R}^3$ be the attention distribution on $\mathrm{I}_m'$, $\mathrm{I}_p'$, and $\mathrm{I}_s'$ at time step $t$, $win$ be the sliding window size employed in the model. Then the processing of words and images are described as follows:

$$c_t = f_v(\tilde{e}_t) \tag{4}$$

where $\tilde{e}_t = [e_{t-win+1}, ..., e_t]$. $c_t$ is the preprocess result of embedding sequence. Function $f_v$ means convolution operation with kernel being $5\times5$, stride being 1, and padding being 2 on word embeddings. After that, attention on three outputs is calculated based on previous hidden state $h_{t-1}$ of Gated Recurrent Unit (GRU) model and current embedding input.

$$a_t = \eta(\mathrm{W}_a \cdot [h_{t-1}, c_t]) \tag{5}$$

$$g_t = \rho(W_g \cdot [c_t, a_t * [I_m^{'}, I_p^{'}, I_s^{'}]]) \quad (6)$$

After that, $g_t$ works as the input of GRU part. Specifically, four operations are calculated. Namely:

$$r_t = \sigma(W_r \cdot [h_{t-1}, g_t])$$
$$z_t = \sigma(W_z \cdot [h_{t-1}, g_t])$$
$$\tilde{h} = \tanh(W_{\tilde{h}} \cdot [r_t * h_{t-1}, g_t])$$
$$h_t = (1 - z_t) * h_{t-1} + z_t * \tilde{h}$$

where $W_r$, $W_z$, and $W_{\tilde{h}}$ are parameters for the input vector $g_t$, reset gate vector $r_t$, update gate vector $z_t$, and output vector $h_t$. $h_t$ participates in the operation of the attention mechanism and GRU operation of next moment. Then the output $y_t$ is calculated through fully-connected operation with parameter $W_y$.

$$y_t = \eta(W_y \cdot h_t)$$

where $y_t \in \mathbb{R}^{D_v}$ is the vector with the same size of $v_t$. The value of each element range is [0, 1]. $y_{ti}$ is the $i$-th element of $y_t$ and $v_{ti}$ is the $i$-th element of $v_t$. Finally, to training our model, the negative log likelihood is employed.

$$l = -\sum_{t}^{N}\sum_{i}^{D_v} v_{ti}\log(y_{ti}) \quad (7)$$

The goal of the model is to minimize the loss $l$ between the true value of verbal explanation and the predicted verbal explanation.

# 5 EXPERIMENT RESULT

In this part, both the verbal explanation of rule-base model and learning model are evaluated. The rule-based model is evaluated by user test. The learning model is evaluated by quantitativel scores.

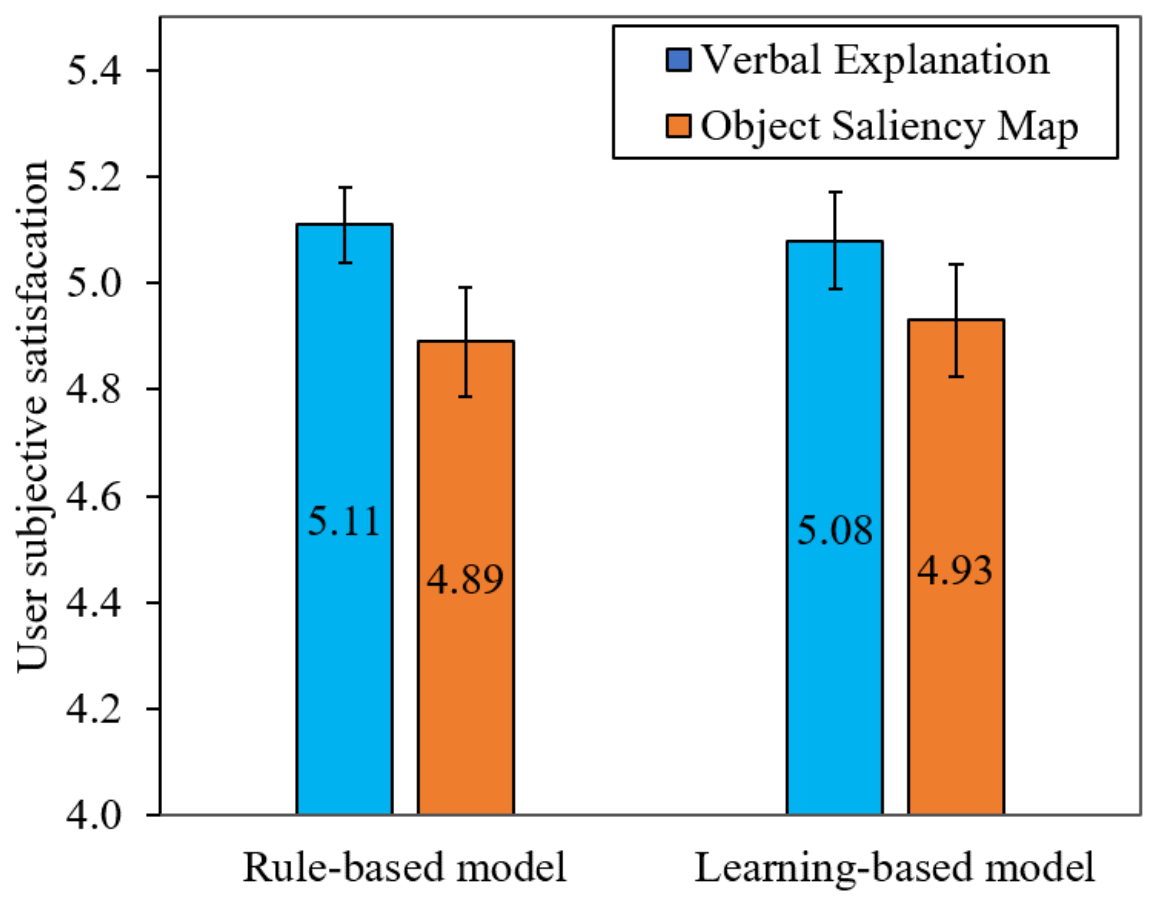

**Figure 4: Subjects satisfaction score.** (5.11 ± 0.07) **and** (4.89 ± 0.09) **for rule-based model group.** (5.08 ± 0.10) **and** (4.93 ± 0.10) **for learning-based model group.**

## 5.1 User test

Online user tests are conducted to validate the appropriateness and effectiveness of both the rule-based and learning-based explanation generation model.

*5.1.1 Questionnaire design.* The online questionnaire consists of an introduction section and two evaluation tasks. The introduction contains basic background knowledge about the Ms. Pacman game, object saliency maps and language generation model. In the two tasks, participants are required to report their subjective satisfactions toward the verbal or object saliency map explanation respectively. Particularly, in each trial of the visual evaluation task, a game screen-shot and the corresponding object saliency map is given. While in the verbal evaluation task, an language explanation generated by either the rule-based or learning based model is given in addition to the two images. The answer in the two tasks scales from 1 (strongly disagree) to 7 (strongly agree).

Participants are required to answer the question "In what degree do you think the saliency map above appropriately explains the current movement of Ms.Pacman?" in a scale from 1 (strongly disagree) to 7 (strongly agree). And the question becomes "In what degree do you think the description above appropriately explains the current movement of Ms.Pacman?"

There are in total 10 trials in each task, which contain randomly selected scenarios from the game episodes played by DRL. The sequence of two tasks and trials in each task is randomized to counterbalance the learning effect. For the rule-based and learning models, tests are conducted separately on different groups of participant to avoid interference. The questionnaire is deployed on Qualtircs.com for public accesses. Participants are recruited through the Amazon Mechanical Turk.

*5.1.2 Result.* In total 150 samples are kept after removing abnormal data. For the rule-based model, the average rating of verbal and visual tasks are 5.11 ± 0.07 (Mean ± Standard Error) and 4.89±0.09, respectively. Paired T-test shows that the rule-based verbal explanations receive significant higher subjective ratings than object saliency maps, $t(74) = 2.989, p = .004$. For the learning model, similar pattern appears that learning-based verbal explanations (5.08 ± 0.10) are better than o-saliency maps (4.93 ± 0.10) in terms of users' satisfaction, $t(74) = 2.020, p = .047$. The result is shown in figure 4. From the result, we can conclude that, the verbal explanation is a more favorable way to interpret the DRL system for game Ms. Pacman.

## 5.2 Validation of Learning model

To ensure the quality of generated verbal explanation from our proposed learning model (named as CNN+ATTEN+GRU), we evaluate the model by comparing with rule-based model and other similar learning models. The training dataset for these models is generated through the object saliency maps and rules introduced earlier. Details are described as follows.

*5.2.1 Dataset.* The dataset we employed to train the model is extracted from 33 episodes of game Ms. Pacman, which automatically played by the DRL system. In practice, less than one quarter of the frames from each episode are valid for the rule-based model as the result of lack of generalization.

**Table 1: Configuration of seven models**

| Method | Input | Encoder | Attention |
|---|---|---|---|
| CNN+GRU | $I_m, I_p, I_s$ | proposed | False |
| Resnet+GRU(state only) | $I_m$ | Resnet | False |
| Resnet+ATTEN+GRU | $I_m, I_p, I_s$ | Resnet | True |
| VGG+GRU(state only) | $I_m$ | VGG | False |
| VGG+ATTEN+GRU | $I_m, I_p, I_s$ | VGG | True |
| CNN+GRU(state only) | $I_m$ | proposed | False |
| CNN+ATTEN+GRU | $I_m, I_p, I_s$ | proposed | True |

Data from 30 episodes are randomly chosen as training data, and the rest 3 episodes are set as testing data. 19419 and 2006 input-output pairs are selected for training and testing data respectively. The ground truth is the result generated through rule-based model.

*5.2.2 Comparison Models.* To evaluate the reliability of proposed learning model, and investigate if the pacman position map, object saliency map, and attention mechanism play their roles, seven models are designed.

The input and attention configuration of models are shown in table 1. $I_m$, $I_p$, $I_s$ means game image, Pacman position map, and object saliency map respectively. 'proposed', 'Resnet', 'VGG' in the encoder column indicate the encoder we introduced in previous section, Resnet network [9], and VGG [28] type. 'True' or 'False' in the 'Attention' column indicates if there is attention mechanism applied. GRU is employed as decoder for all the seven models.

In the seven models, only the image encoder and the attention mechanism are changed. In models CNN+GRU, CNN+GRU (state only), and CNN+ATTEN+GRU, the proposed image encoder operations are designed. In models VGG+GRU (state only) and VGG+ATTEN+GRU, VGG network [28] is employed to extract image features. In models Resnet+GRU (state only) and Resnet+ATTEN+GRU, Resnet network [9] is employed to process inputs images.

The game image, Pacman position map, and object saliency maps act evenly in model CNN+GRU, because there is no attention mechanism in this model. Models which are marked as '(state only)' only take game image as encoder input. Three models, tailed with '(state only)', are designed to see if Pacman position map and object saliency maps contribute or not. In models Resnet+ATTEN+GRU, VGG+ATTEN+GRU, and CNN+ATTEN+GRU, dynamically calculated attention on encoder output is implemented.

*5.2.3 Model Performance.* We compare the performance of the seven models through the training loss and the BLEU score, which is shown in figure 5 and table 2 respectively. The training loss reveals the deviation between model output and the given ground-truth. BLEU score quantitatively measure the similarity between the generated explanations and rule-based result.

As shown in both metrics, the proposed model#7 CNN+ATTEN+GRU outperforms other six models. It can be seen that CNN+ATTEN+GRU get the highest score on BLEU-1 (0.714), BLEU-2 (0.631), BLEU-3 (0.560), and BLEU-4 (0.501) on testing data. We further analyze the result by probing into the contribution of the input data, attention mechanism, and encoder structures respectively.

*Input channels and attention mechanism:* Models with attention and employing three kinds of images as input perform better than their pairs which only take game image as input. Namely,

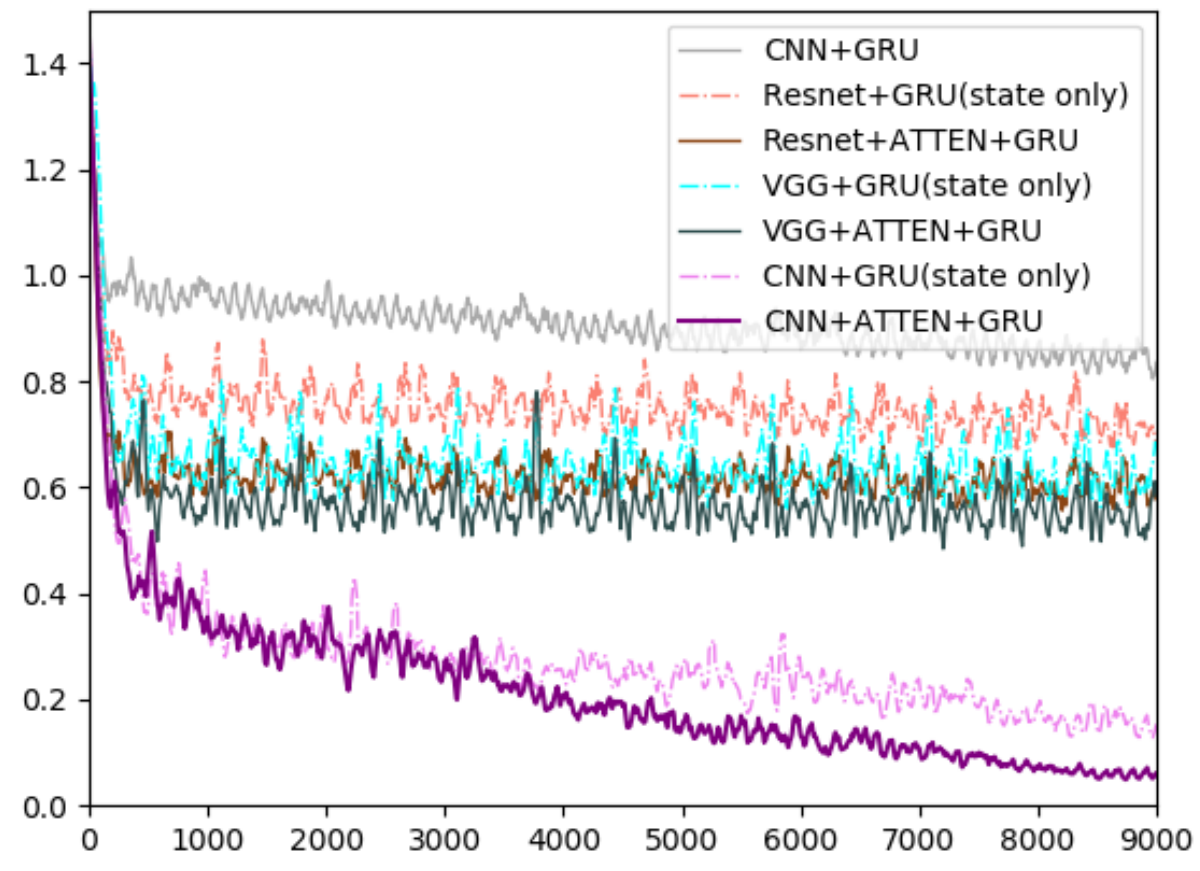

**Figure 5: The loss of seven different models during the training process. The x-axis is the training step, y-axis is value of loss.**

**Table 2: BLEU score on training and testing data, Model#1 to Model#7 are CNN+GRU, Resnet+GRU(state only), Resnet+ATTEN+GRU, VGG+GRU(state only), VGG+ATTEN+GRU, CNN+GRU(state only), CNN+ATTEN+GRU respectively**

| **BLEU Socre on training data** | | | | |
|---|---|---|---|---|
| **Method** | **BLEU-1** | **BLEU-2** | **BLEU-3** | **BLEU-4** |
| Model#1 | 0.481 | 0.323 | 0.230 | 0.153 |
| Model#2 | 0.536 | 0.399 | 0.293 | 0.209 |
| Model#3 | 0.633 | 0.514 | 0.422 | 0.344 |
| Model#4 | 0.621 | 0.478 | 0.382 | 0.304 |
| Model#5 | 0.587 | 0.469 | 0.380 | 0.306 |
| Model#6 | 0.797 | 0.735 | 0.681 | 0.635 |
| Model#7 | **0.910** | **0.883** | **0.857** | **0.834** |
| **BLEU Socre on testting data** | | | | |
| **Method** | **BLEU-1** | **BLEU-2** | **BLEU-3** | **BLEU-4** |
| Model#1 | 0.493 | 0.337 | 0.244 | 0.166 |
| Model#2 | 0.544 | 0.414 | 0.309 | 0.224 |
| Model#3 | 0.638 | 0.519 | 0.428 | 0.350 |
| Model#4 | 0.626 | 0.490 | 0.396 | 0.319 |
| Model#5 | 0.596 | 0.484 | 0.396 | 0.322 |
| Model#6 | 0.703 | 0.620 | 0.548 | 0.487 |
| **Model#7** | **0.714** | **0.631** | **0.560** | **0.501** |

Resnet+ATTEN+GRU performs better than Resnet+GRU (state only), VGG+ATTEN+GRU is better than VGG+GRU (state only), and CNN+ATTEN+GRU is better than CNN+GRU (state only) on higher BLEU. A more straightforward visual of the attention mechanism is shown as figure 6, which presents the attention distribution on three kinds of input from CNN+ATTEN+GRU. Lighter colors refer to high amount of attention allocation. It can be seen that the attention distribution changes during verbal generation process. It means, game image, Pacman position map and object saliency maps contribute differently to different components in the generated explanation. There is one interesting phenomenon that Pacman position map get the most attention during the generation process as is shown in the figure. The reason for that may be that Pacman is the main agent when playing the game.

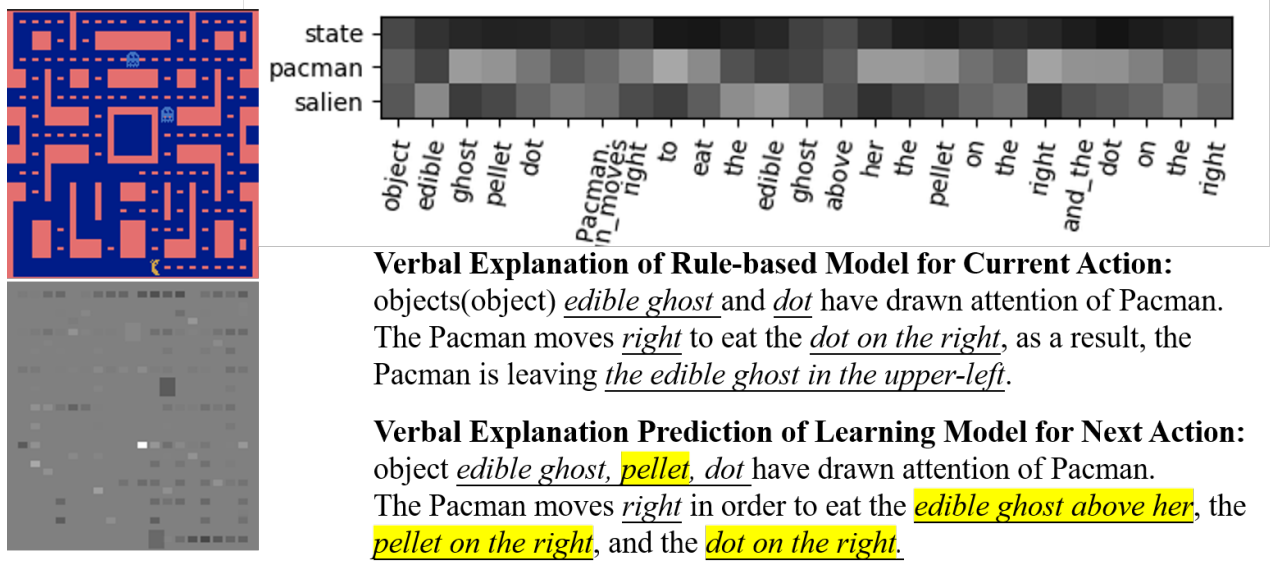

**Figure 6: The attention weight distribution on game image, Pacman position image and object saliency map from model CNN+ATTEN+GRU. Light rectangle means high attention. Dark rectangle means low attention.**

*Encoder structures:* In these models, the proposed encoder performs better than Resnet and VGG network, even though Resnet and VGG are proven to be powerful in other scenarios. The reason might be that the game images of Ms. Pacman are highly structural similar with fixed background and objects, which are different from real life pictures, such as images in ImageNet. The proposed big convolution kernel with max-pooling contributes to extracting distinguishable features from similar images.

#### 5.2.4 *Action Prediction Evaluation.*

The learning model is able to generate explanations for both the current and future game state. We evaluate the effectiveness of prediction by comparing the predicted action in explanations and the real game action in future states. The action prediction evaluation is implemented on model CNN+GRU(state only) and model CNN+ATTEN+GRU, which have relatively better performance. Recall and precision on each action are calculated. CNN+ATTEN+GRU also shows better performance, as is shown in table 3. 'action-r' and 'action-p' means the recall and precision separately. The overall accuracy is also given.The accuracy on testing data is higher than 0.64 on both models.

### 5.3 Explanations from both models

As we mentioned above, the rule-based model construct explanations for strongly constrained situations under a series of objective and rationalized rules. The learning model mines the implicit logic of generating verbal explanation with the help of rule-based model to all the situations. The rule-based model supports learning model by providing training data. In return, the learning model generalizes the rule-based model to all the situations, including the ones beyond the capabilities of the rule-based model. Moreover, the learning model is capable of generating explanation for future possible actions.

In this section, four cases are shown in table.4 from both rule-based model and learning model (CNN+ATTEN+GRU). The first two cases show the explanations from both models for Pacman's current action. The last two cases show the explanations from rule-based model for Pacman's current action and explanations from model CNN+ATTEN+GRU for Pacman's next action.

## 6 CONCLUSION

This paper present two models working cooperatively on generating objective verbal explanations for DRL systems.

We firstly present a rule-based model which generates the verbal explanations for strongly constrained situations. This model works with high accuracy, but lack of generalization. Experiments show that explaining DRL system with natural language generated by the rule-based model gains higher user satisfaction than that with only object saliency map, showing the usefulness of verbal explanations.

Then, we propose a learning model, which beyond the capabilities of the static rule-based model, to expand the rule-based model to general situations. Seven variants of the learning model are designed to ensure the quality of generated explanations. Experiment results show that object saliency map, the proposed encoder, and attention mechanism contribute to the verbal generation process.

This paper provide clues for verbal explanation of DRL systems by employing the application of Atari game Ms. Pacman. We hope our work throw light on explaining increasingly complex artificial intelligent models.

**Table 3: Evaluation of Pacman Action Prediction**

| **Training Data** | **up-r** | **up-p** | **down-r** | **down-p** | **left-r** | **left-p** | **right-r** | **right-p** | **accuracy** |
|---|---|---|---|---|---|---|---|---|---|
| CNN+GRU(state only) | 0.7593 | 0.7816 | 0.5841 | 0.7667 | 0.7499 | 0.7131 | 0.7332 | 0.6726 | 0.7198 |
| CNN+ATTEN+GRU | 0.9353 | 0.9244 | 0.7913 | 0.9535 | 0.9701 | 0.8720 | 0.9194 | 0.9514 | 0.9181 |
| **Testing Data** | **up-r** | **up-p** | **down-r** | **down-p** | **left-r** | **left-p** | **right-r** | **right-p** | **accuracy** |
| CNN+GRU(state only) | 0.7149 | 0.7149 | 0.4808 | 0.6318 | 0.6194 | 0.6613 | 0.7104 | 0.5869 | 0.6426 |
| CNN+ATTEN+GRU | 0.7472 | 0.7390 | 0.5781 | 0.6873 | 0.7045 | 0.6299 | 0.645 | 0.6649 | 0.6735 |

**Table 4: Cases of generated verbal explanation from both rule-based model and learning model (CNN+ATTEN+GRU)**

| **Game Image** | **Saliency Map (Image Explanation)** | **Verbal Generation Result (Verbal Explanation)** |
|---|---|---|
| 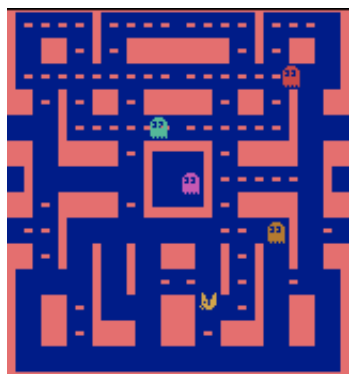 | 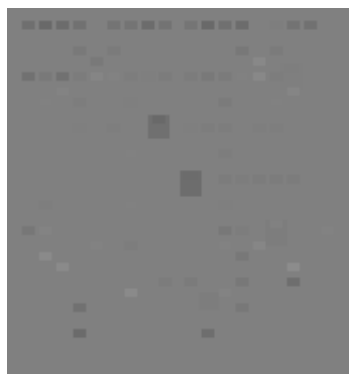 | ***Verbal Explanation of Rule-based Model for Current Action:***<br>objects(object) ghost and dot have drawn attention of pacman.<br>the pacman moves up to eat the dot on the left.<br>as a result, the pacman is approaching the 2 ghosts<br>in the upper-right and in the upper-left.<br>***Verbal Explanation of Learning Model for Current Action:***<br>objects(object) ghost and dot have drawn attention of pacman.<br>the pacman moves left to eat the dot on the left.<br>as a result, the pacman is approaching the 2 ghosts above<br>him and in the upper-right. |
| 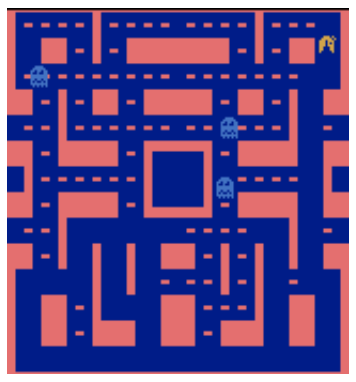 | 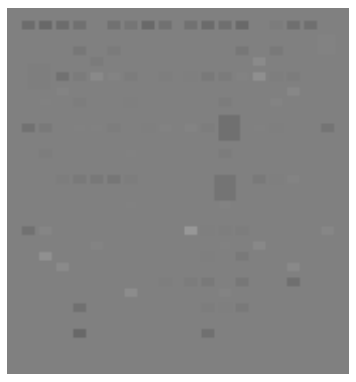 | ***Verbal Explanation of Rule-based Model for Current Action:***<br>the pacman is moving down.<br>because he wants to eat the edible ghost in the lower-left.<br>***Verbal Explanation of Learning Model for Current Action:***<br>object edible ghost dot have drawn attention of pacman.<br>the pacman is moving down.<br>because he wants to eat the 2 edible ghosts in the lower-left and the dot below him. |
| 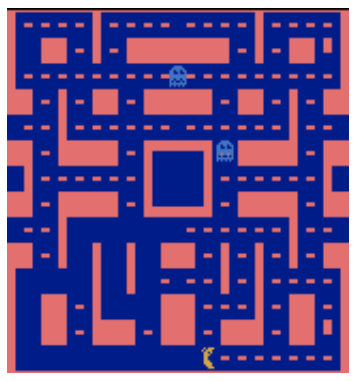 | 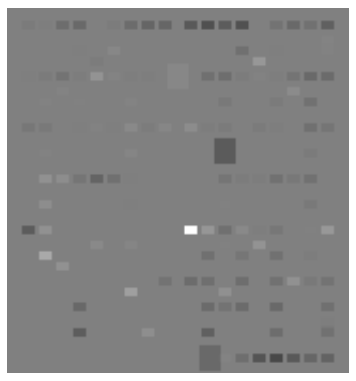 | ***Verbal Explanation of Rule-based Model for Current Action:***<br>objects(object) edible ghost and dot have drawn attention of pacman.<br>the pacman moves right to eat the dot on the right.<br>as a result, the pacman is leaving the edible ghost in the upper-left.<br>***Verbal Explanation Prediction of Learning Model for Next Action:***<br>object edible ghost pellet dot have drawn attention of pacman.<br>the pacman moves right to eat the edible ghost above her the pellet<br>on the right and the dot on the right. |
| 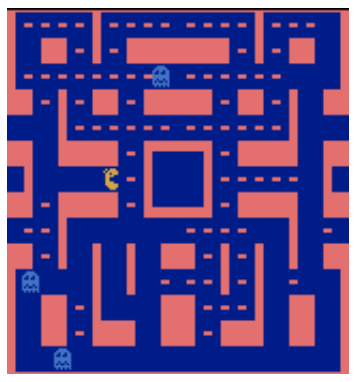 | 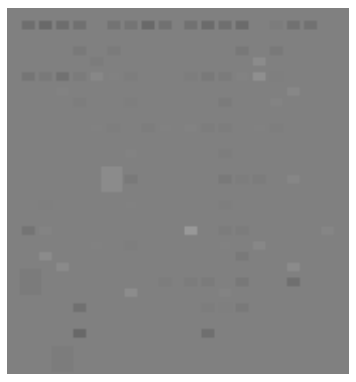 | ***Verbal Explanation of Rule-based Model for Current Action:***<br>objects(object) edible ghost pellet and dot have drawn attention of pacman.<br>the pacman moves left to eat the 2 edible ghosts in the lower-left and below her,<br>the pellet in the upper-left and the dot below him.<br>as a result, the pacman is leaving the edible ghost in the upper-right.<br>***Verbal Explanation Prediction of Learning Model for Next Action:***<br>objects(object) edible ghost and dot have drawn attention of pacman.<br>the pacman moves right to eat the edible ghost in the upper-right and<br>the dot in the upper-right.<br>as a result, the pacman is leaving the 2 edible ghosts below him and in the lower-left. |